# The 2017 DAVIS Challenge on Video Object Segmentation

Jordi Pont-Tuset, Federico Perazzi, Sergi Caelles,
Pablo Arbeláez, Alexander Sorkine-Hornung, and Luc Van Gool

**Abstract**—We present the *2017 DAVIS Challenge on Video Object Segmentation*, a public dataset, benchmark, and competition specifically designed for the task of video object segmentation. Following the footsteps of other successful initiatives, such as ILSVRC [1] and PASCAL VOC [2], which established the avenue of research in the fields of scene classification and semantic segmentation, the DAVIS Challenge comprises a dataset, an evaluation methodology, and a public competition with a dedicated workshop co-located with CVPR 2017. The DAVIS Challenge follows up on the recent publication of DAVIS (Densely-Annotated VIdeo Segmentation [3]), which has fostered the development of several novel state-of-the-art video object segmentation techniques. In this paper we describe the scope of the benchmark, highlight the main characteristics of the dataset, define the evaluation metrics of the competition, and present a detailed analysis of the results of the participants to the challenge.

✦

## 1 INTRODUCTION

Public benchmarks and challenges have been an important driving force in the computer vision field, with examples such as Imagenet [1] for scene classification and object detection, PASCAL [2] for semantic and object instance segmentation, or MS-COCO [4] for image captioning and object instance segmentation. From the perspective of the availability of annotated data, all these initiatives were a boon for machine learning researchers, enabling the development of new algorithms that had not been possible before. Their challenge and competition side motivated more researchers to participate and push towards the new different goals, by setting up a fair environment where test data are not publicly available.

The Densely-Annotated VIdeo Segmentation (DAVIS) initiative [3] provided a new dataset with 50 high-definition sequences with all their frames annotated with object masks at pixel-level accuracy, which has allowed the appearance of a new breed of video object segmentation algorithms [5], [6], [7], [8] that pushed the quality of the results significantly, almost getting to the point of saturation of the original dataset (around 80% performance by [5] and [6]). We will refer to this version of the dataset as DAVIS 2016.

To further push the performance in video object segmentation, we present *the 2017 DAVIS Challenge on Video Object Segmentation*, which consists of a new, larger, more challenging dataset (which we refer to as DAVIS 2017) and

- J. Pont-Tuset, S. Caelles, and L. Van Gool are with the Computer Vision Laboratory, ETH Zürich, Switzerland.
- F. Perazzi and A. Sorkine-Hornung are with Disney Research, Zürich, Switzerland.
- P. Arbeláez is with the Department of Biomedical Engineering, Universidad de los Andes, Colombia.

Contacts and updated information can be found in the challenge website: http://davischallenge.org

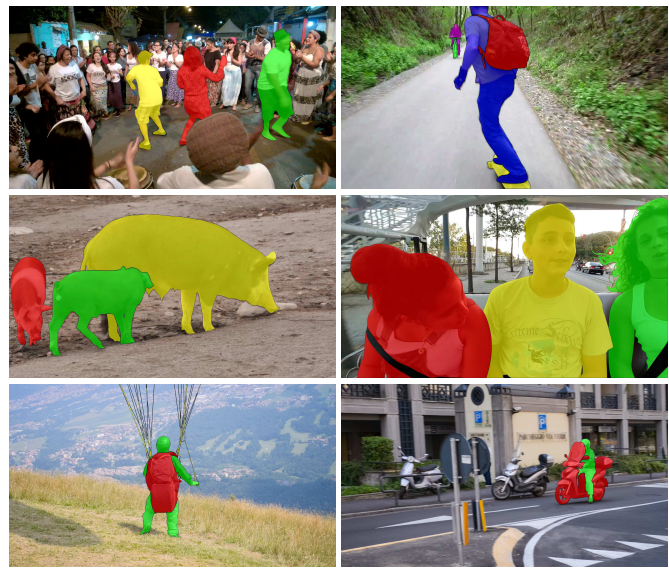

Fig. 1. Example annotations of the DAVIS 2017 dataset: The four first images come from new videos, the last two from videos originally in the DAVIS dataset re-annotated with multiple objects.

a public challenge competition and workshop. As the main new challenge, the new sequences have more than one annotated object in the scene, and we have re-annotated the original ones that have more than one visible object. The complexity of the videos has also increased with more distractors, smaller objects and fine structures, more occlusions and fast motion, etc. Overall, the new dataset consists of 150 sequences, totaling 10459 annotated frames and 376 objects. Figure 1 shows a set of example frames with the corresponding overlaid object annotations, and Section 2 gives detailed facts and figures about the new annotations.

We hosted a public competition challenge based on the task of segmenting the objects in the video given the detailed segmentation of each object in the first frame (the so-called semi-supervised scenario in DAVIS 2016). Section 3 describes the task in detail and the metrics used to evaluate the results. The results of the competition were presented in a workshop in the Computer Vision and Pattern Recognition (CVPR) conference 2017, in Honolulu, Hawaii. The challenge received entries from 22 different teams and brought an improvement of 20% to the state of the art. Section 4 gives a detailed analysis of the results of the top-performing teams.

|  | DAVIS 2016 | | | DAVIS 2017 | | | | |
|---|---|---|---|---|---|---|---|---|
|  | `train` | `val` | **Total** | `train` | `val` | `test-dev` | `test-challenge` | **Total** |
| Number of sequences | 30 | 20 | **50** | 60 | 30 | 30 | 30 | **150** |
| Number of frames | 2079 | 1376 | **3455** | 4219 | 2023 | 2037 | 2180 | **10459** |
| Mean number of frames per sequence | 69.3 | 68.8 | **69.1** | 70.3 | 67.4 | 67.9 | 72.7 | **69.7** |
| Number of objects | 30 | 20 | **50** | 138 | 59 | 89 | 90 | **376** |
| Mean number of objects per sequence | 1 | 1 | **1** | 2.30 | 1.97 | 2.97 | 3.00 | **2.51** |

TABLE 1
Size of the DAVIS 2016 and 2017 dataset splits: number of sequences, frames, and annotated objects.

## 2 DATASET FACTS AND FIGURES

The main new challenge added to the DAVIS sequences in its edition of 2017 is the presence of multiple objects in the scene. As it is well known, the definition of an object is granular, as one can consider a person as including the trousers and shirt, or consider them as different objects. In DAVIS 2016 the segmented object was defined as the main *object* in the scene with a distinctive motion. In DAVIS 2017, we also segment the main moving objects in the scene, but we also divide them by semantics, even though they might have the same motion. Specifically, we generally segmented people and animals as a single instance, together with their clothes, (including helmet, cap, etc.), and separated any object that is carried and easily separated (such as bags, skis, skateboards, poles, etc.). As an example, Figure 2 shows different pairs of DAVIS 2016 segmentation (left) together with their DAVIS 2017 multiple-object segmentations.

As is a common practice in the computer vision challenges, we divide our dataset into different splits. First of all, we extend the `train` and `val` sets of the original DAVIS 2016, with annotations that are made public for the whole sequence. We then define two other test sets (`test-dev` and `test-challenge`), for which only the masks on the first frames are made public. We set up an evaluation server in Codalab where researchers are able to submit their results, download an evaluation file, and publish their performance on the public leaderboard. During the challenge, submissions for `test-dev` were unlimited and for a longer period of time, whereas `test-challenge`, which determined the winners, was only open for a short period of time and for a limited number of submissions. After the challenge, `test-dev` submissions remain open as we expect future methods to publish their results on this set.

Table 1 shows the number of sequences, frames, and objects on each of the dataset splits. Please note that `train` and `val` in DAVIS 2017 include the sequences of the respective sets in DAVIS 2016 with multiple objects annotated when applies. This is the reason why the mean number of objects per sequence is smaller in these two sets, despite all new sequences have around 3 objects per sequence in mean. The length of the sequences is kept similar to DAVIS 2016: around 70 frames.

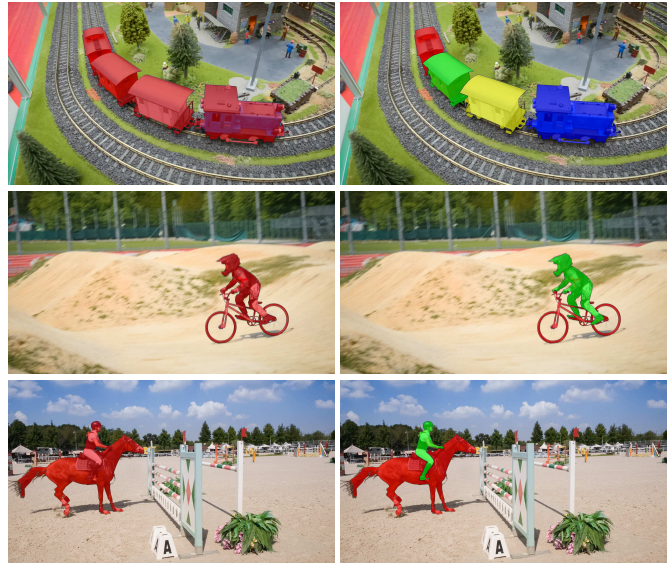

Fig. 2. Example annotations of the DAVIS 2017 vs the single-object counterpart in DAVIS 2016: Semantics play a role even if the objects have the same motion.

In terms of resolution, the majority of new sequences are at 4k resolution (3840×2160 pixels), but there are also some 1440p, 1080p, and 720p images at its raw resolution. Despite this, the challenge will be on the downsampled 480p images, as it was the *de facto* standard for DAVIS 2016, and to facilitate their processing given the large amount of frames. We plan to increase the resolution used in future editions of the challenge.

## 3 TASK DEFINITION AND EVALUATION METRICS

The challenge will be focused on the so-called *semi-supervised* video object segmentation task, that is, the algorithm is given a video sequence and the mask of the objects in the first frame, and the output should be the masks of those objects in the rest of the frames. This excludes more supervised approaches that include a human in the loop (interactive segmentation) and unsupervised techniques (no initial mask is given). Please note that all objects in a frame have its unique identifier and so the expected output is a set of indexed masks by identifier.

|  | Measure | 1st [9] | 2nd [10] | 3rd [11] | 4th [12] | 5th [13] | 6th [14] | 7th [15] | 8th [16] | 9th [17] | OSVOS [5] |
|---|---|---|---|---|---|---|---|---|---|---|---|
| $\mathcal{J}\&\mathcal{F}$ | Mean $\mathcal{M}$ ↑ | **69.9** | 67.8 | 63.8 | 61.5 | 57.7 | 56.9 | 53.9 | 50.9 | 49.7 | 49.0 |
| $\mathcal{J}$ | Mean $\mathcal{M}$ ↑ | **67.9** | 65.1 | 61.5 | 59.8 | 54.8 | 53.6 | 50.7 | 49.0 | 46.0 | 45.6 |
|  | Recall $\mathcal{O}$ ↑ | **74.6** | 72.5 | 68.6 | 71.0 | 60.8 | 59.5 | 54.9 | 55.1 | 49.3 | 46.7 |
|  | Decay $\mathcal{D}$ ↓ | 22.5 | 27.7 | **17.1** | 21.9 | 31.0 | 25.3 | 32.5 | 21.3 | 33.1 | 34.2 |
| $\mathcal{F}$ | Mean $\mathcal{M}$ ↑ | **71.9** | 70.6 | 66.2 | 63.2 | 60.5 | 60.2 | 57.1 | 52.8 | 53.3 | 52.5 |
|  | Recall $\mathcal{O}$ ↑ | 79.1 | **79.8** | 79.0 | 74.6 | 67.2 | 67.9 | 63.2 | 58.3 | 58.4 | 56.0 |
|  | Decay $\mathcal{D}$ ↓ | 24.1 | 30.2 | **17.6** | 23.7 | 34.7 | 27.6 | 33.7 | 23.7 | 36.4 | 35.9 |
| $\mathcal{T}$ | Mean $\mathcal{M}$ ↓ | **26.0** | 37.1 | 41.6 | 54.3 | 51.3 | 67.1 | 61.3 | 75.6 | 65.9 | 67.1 |

TABLE 2
**Final results of the 2017 DAVIS Challenge on the Test-Challenge set:** As a baseline, we add the naive generalization of OSVOS to multiple objects by tracking each of them independently.

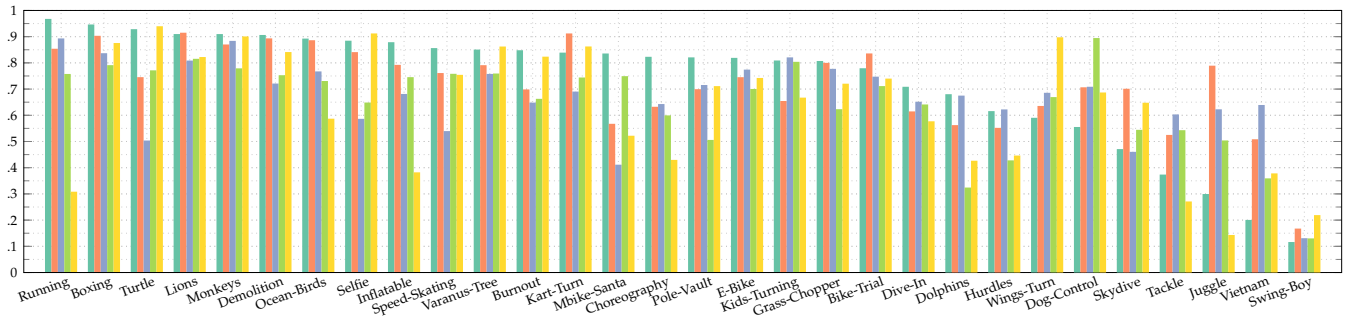

Fig. 3. **Per-sequence results of mean region similarity and contour accuracy ($\mathcal{J}\&\mathcal{F}$):** A random object from each sequence is selected.

Given a mask of a specific object given by an algorithm and the ground-truth mask of that same object in a specific frame, we use the region ($\mathcal{J}$) and boundary ($\mathcal{F}$) measures proposed in DAVIS 2016 [3]. Specifically, the former computes the number of pixels of the intersection between the two masks and divides it by the size of the union (also called Intersection over Union - IoU, or Jaccard index). The latter evaluates the accuracy in the boundaries, via a bipartite matching between the boundary pixels of both masks. The final boundary measure is the F measure between the precision and recall of the matching. Please refer to [3] for further description and discussion about these measures.

As of this edition, we discard the temporal instability ($\mathcal{T}$) given that its behavior is very affected by heavy occlusions. In DAVIS 2016 we computed the measures on the subset of sequences with less occlusions but in DAVIS 2017 occlusions happen much more often, which would make the results less significant. Despite this, we encourage researchers to keep evaluating $\mathcal{T}$ and reporting it in the papers on the subset of selected sequences (available in the official code), since it is informative of the stability of the results.

As an overall measure of the performance of each algorithm we will compute the mean of the measures ($\mathcal{J}$ and $\mathcal{F}$) over all object instances. Formally, let $S$ be a set of sequences, and $O_S$ the set of annotated objects in these sequences. Given an object $o \in O_S$, $s(o) \in S$ is the sequence where the given object appears. Then, let $F_s$ be the set of frames in sequence $s \in S$. Given a metric $\mathcal{M}$, the mean performance metric $m(\mathcal{M}, S)$ in the sequence set $S$ is then defined as:

$$m(\mathcal{M}, S) = \frac{1}{|O_S|} \sum_{o \in O_S} \frac{1}{|F_{s(o)}|} \sum_{f \in F_{s(o)}} \mathcal{M}(m_o^f, g_o^f)$$

where $m_o^f$ and $g_o^f$ are the binary masks of the object and ground truth, respectively, of object $o$ in frame $f$.

The overall performance metric that defines the ranking in a given set of the challenge is defined as:

$$M(S) = \frac{1}{2}\left[m(\mathcal{J}, S) + m(\mathcal{F}, S)\right]$$

as the average of the mean region and contour accuracies.

The performance of the metric in a given sequence $s \in S$ is defined as $m(\mathcal{M}, \{s\})$. Please note that we will report the metric per sequence as an informative measure, but the overall metric will not be the mean of the per-sequence values but per object as defined above, that is, in general $M(S) \neq \sum_{s \in S} M(\{s\})$.

## 4 ANALYSIS OF THE RESULTS

**Overview of the techniques presented**: The winner technique [9] combines MaskTrack, the mask propagation network presented in [6], with a re-identification module that recovers the object when the mask propagation fails. The second technique [10], closely following the winner, extends the training of MaskTrack [6] with different variations of the first segmentation by means of a *lucid dreaming* module. The rest of techniques present improvements over OS-VOS [5], an appearance model trained on the first frame [13], [15], [16], or make use of instance re-identification modules (as the winner method) [11], object proposals [12], [17], or spatial propagation networks [14].

**Global statistics**: Let us start with a global overview of the evaluation results of the 2017 DAVIS challenge. Table 2 shows the results from the accepted participants, using the metrics defined in the original DAVIS, plus the mean over $\mathcal{J}\&\mathcal{F}$, which defined the challenge winner. We also add OSVOS [5] as baseline: each object of the sequence segmented independently and merged together naively.

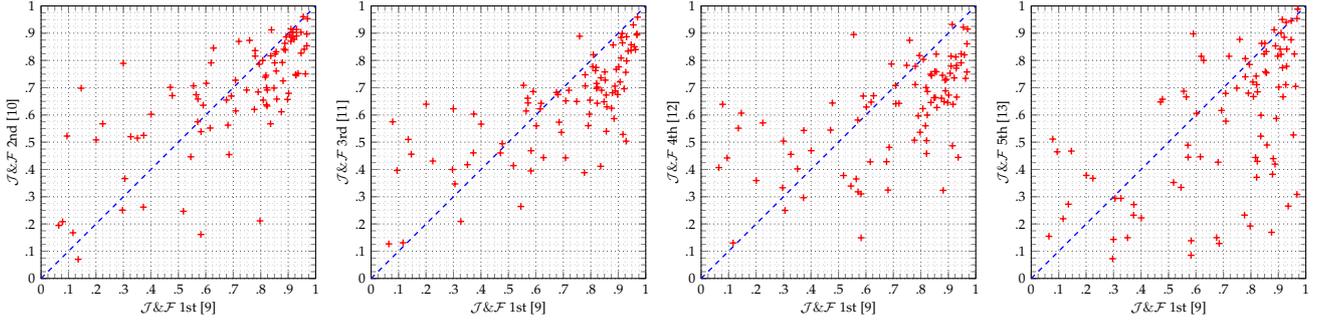

Fig. 4. **Performance comparison:** (Mean $\mathcal{J}\&\mathcal{F}$) between the first methods, for the whole set of objects of DAVIS-2017 Test-Challenge.

| Measure | | 1st [9] | 2nd [10] | 3rd [11] | 4th [12] | 5th [13] | 6th [14] | 7th [15] | 8th [16] | 9th [17] | OSVOS [5] |
|---|---|---|---|---|---|---|---|---|---|---|---|
| $\mathcal{J}\&\mathcal{F}$ | Mean $\mathcal{M}\uparrow$ | **82.4** | 81.9 | 75.8 | 74.3 | 76.8 | 78.0 | 77.3 | 63.8 | 64.8 | 67.2 |

TABLE 3
**Results evaluated as a single foreground-background classification:** all segmented objects as foreground, rest background.

We can observe that the nine methods cover the whole spectrum between barely improving over the OSVOS baseline (49.7%) and being 20 points above it (69.9%). The winner is 2.1% over the runner up, and wins not only in terms of $\mathcal{J}\&\mathcal{F}$ but also in four other metrics.

Figure 3 breaks down the performance into sequences, by showing the result in one random object from each of them. The sequences are sorted by the performance of the winner method on each of them. We can observe some easy objects where all techniques perform well (Boxing), others challenging for everyone (Swing-Boy), and others with a large variance between techniques (Juggle).

To further explore whether there are some trends in the performance in each sequence, Figure 4 shows the compared performance on all 92 objects between the first technique and the following ones. Although some correlation can be seen, there are still a significant amount of objects that are well segmented by one of the techniques and not by the other, proof of the variety of techniques presented, and that there is still room for improvement. To further make this point, the oracle result obtained from choosing the best result for each object from a set of techniques is 75.3%, 76.7%, 77.4%, and, 77.9%; when combining up to the second, third, fourth, and fifth best techniques, respectively. In other words, combining the first two gets a +5.4% boost over the first technique.

**Failure analysis**: To further analyze where the current methods fail, and to isolate how much having multiple objects adds to the challenge, Table 3 shows the evaluation of the same results but transforming all object masks (and ground truth) to a single foreground-background segmentation. We can observe that the results increase dramatically (e.g. up to 82.4% from 69.9% in the case of the winner).

One of the thesis to justify this behavior is that current methods are good at foreground-background segmentation but struggle to separate between foreground objects. We investigate into it by separating the set of misclassified pixels into global false positives, global false negatives, and identity switches (pixel label swapped between objects). Linking to the previous experiment, the first two type of errors are the ones that are also done in the single-object case, whereas the latter is considered correct. False positives are further divided into FP-close and FP-far, to separate between boundary inaccuracies and spurious detections (as done in [5]). Figure 5 depicts the percentage of pixels in the image that fall into each of these categories. Surprisingly, false negatives dominate the types of errors, not only in front of the false positives (as also observed in [5]), but also with respect to identity switches. We believe a possible explanation of this imbalance is that spurious detections (FP-Far) might be more noticeable to humans, so in the process of manually adjusting the parameters of an algorithm, the solutions leading to higher false negatives are unconsciously chosen.

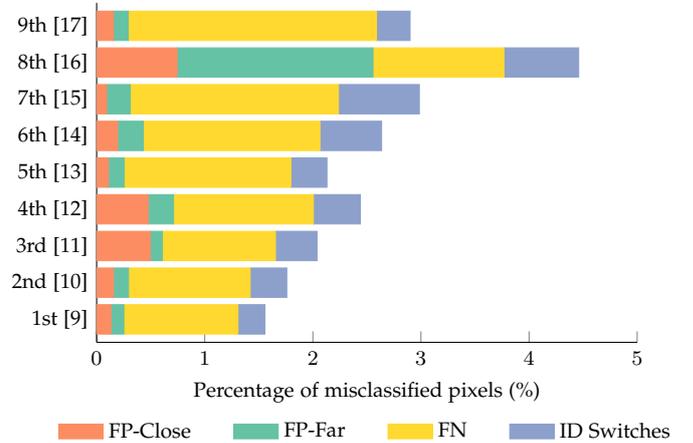

Fig. 5. **Classification of the error pixels into false positives, false negatives, and swapped object:** Errors are given in percentage of erroneous pixels in the image, averaged per object.

Note that the percentage of misclassified pixels is not directly comparable to $\mathcal{J}$ and $\mathcal{F}$ (5th has less wrongly classified pixels than 4th), because the latter depend also on the size of each object. Let us therefore analyze the performance of the methods with respect to the size of the objects: a foreground object that is completely missed but it

5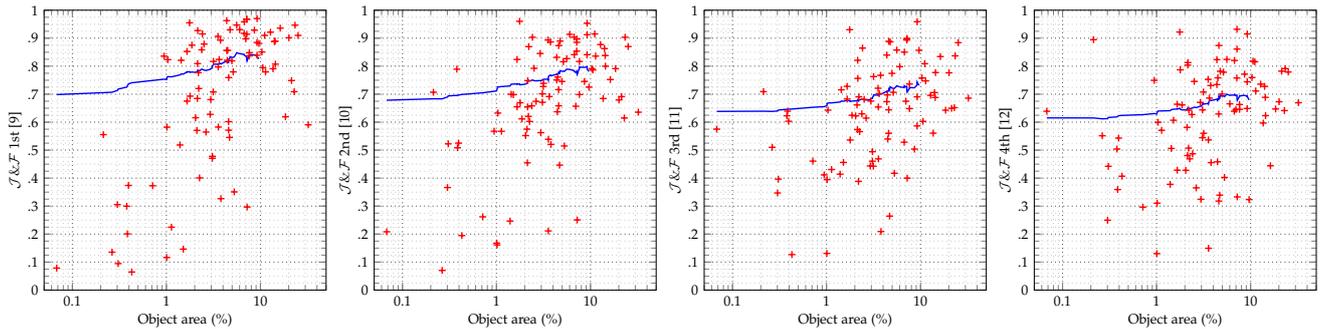

Fig. 6. **Performance on all 92 objects (+) with respect to their area, for the first four techniques:** The line (———) shows the mean quality when we remove the objects smaller than a certain size.

is very small would not affect the foreground-background result significantly despite affecting the multi-object case significantly. To analyze this, Figure 6 shows the performance of the first four methods in all objects with respect to their area (red crosses +).

We can observe that there is indeed a certain tendency for small objects to be more difficult than the large ones. The blue lines (———) depict the evolution of the mean quality when removing the smaller objects than a certain threshold. This way, the curve starts at the mean over all objects (challenge result), and each step removes the smallest object from the mean. We can observe that in some methods such as the winner, the quality goes up to 85% (+15%) if we remove all objects whose area is below 5% of the image area. The fourth method, in contrast, is more robust to object size and gets only up to +8%.

To sum up, multiple objects add a new challenge of identity preserving, and although it is significant, the errors are still dominated by false negatives, as in the single-object case. The presence of small objects also poses a new challenge that was not observed in the single-object case simply because the sample of objects did not include small ones.

**Qualitative results**: Figure 7 shows a set of selected qualitative results comparing the performance of the first four method (right-most four columns) and the ground truth (left-most column), covering a wide range of challenges and behaviors. Please refer also to Figure 3 to match each sequence to its mean quantitative performance. On the first row, an *easy* sequence in which performance is good on all methods. The three following rows (turtle, mbike-santa, and choreography) show cases where the winner indeed obtains the best results, despite the identity switch in one of the choreographers. Rows 5 (Vietnam) and 6 (juggle) show the opposite: the runner up showing better performance than the winner. In the first case there is again an identity switch, whereas in the second the ball is completely lost. The final row (swing-boy) shows a sequence where none of the techniques correctly segment the chain of the swing.

## 5 CONCLUSIONS

This paper presented the *2017 DAVIS Challenge on Video Object Segmentation*, which extends the DAVIS (Densely-Annotated VIdeo Segmentation [3]) dataset with more videos and more challenging scenarios (especially more than one object per sequence). We also defined the metrics and competitions of the public challenge that we hosted in conjunction with CVPR 2017, and provided a detailed analysis of the results obtained by the participating teams.

### ACKNOWLEDGMENTS

Research partially funded by the workshop sponsors: Google, Disney Research, NVIDIA, Prof. Luc Van Gool's Computer Vision Lab at ETHZ, and Prof. Fuxin Li's group at the Oregon State University. The authors gratefully acknowledge support by armasuisse, and thank NVIDIA Corporation for donating the GPUs used in this project.

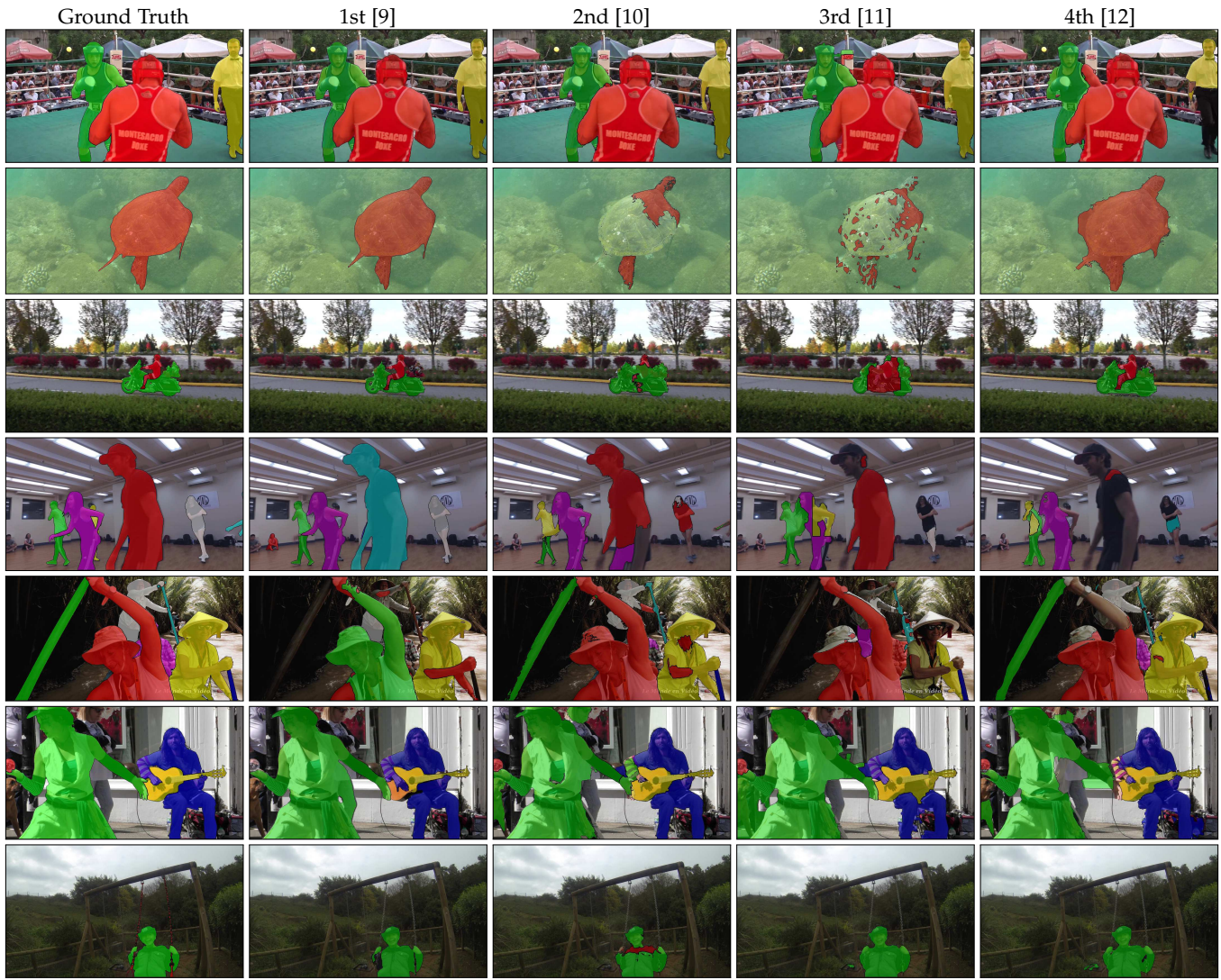

Fig. 7. **Qualitative results**: Comparison between top-performing methods in different sequences.